\pdfoutput=1
\documentclass[11pt]{article}
\usepackage[]{acl}
\usepackage[T1]{fontenc}
\usepackage[utf8]{inputenc}
\usepackage{microtype}
\usepackage{newtxtext}
\usepackage{newtxmath}
\usepackage{amsmath}
\usepackage{booktabs}
\usepackage{tabularx}
\usepackage{graphicx}
\usepackage{subcaption}

\title{%
    Revisiting the Phenomenon of Syntactic Complexity Convergence\\
    on German Dialogue Data}

\author{%
    Yu Wang \and Hendrik Buschmeier\\
        Digital Linguistics Lab, Department of Linguistics\\
        Faculty of Linguistics and Literary Studies\\
        Bielefeld University, Bielefeld, Germany\\
        \texttt{\{y.wang,hbuschme\}@uni-bielefeld.de}
}

\begin{document}
\maketitle

\begin{abstract}
We revisit the phenomenon of syntactic complexity convergence in conversational interaction, originally found for English dialogue, which has theoretical implication for dialogical concepts such as mutual understanding. We use a modified metric to quantify syntactic complexity based on dependency parsing. The
results show that syntactic complexity convergence can be statistically confirmed in one of three selected German datasets that were analysed.
Given that the dataset which shows such convergence is much larger than the other two selected datasets, the empirical results indicate a certain degree of linguistic generality of syntactic complexity convergence in conversational interaction.
We also found a different type of syntactic complexity convergence in one of the datasets while further investigation is still necessary.
\end{abstract}

\section{Introduction}
\label{sec:introduction}

The interactive alignment theory \citep{pickering2004interactive} states that, in interaction, mutual understanding is reached through the support of adaptive processes, which result in a reduction of the communicative efforts of the dialogue participants.
\citet{pickering2004interactive} have mentioned the co-adaptivity of interlocutors' verbal behaviour on the following six levels: phonetic, phonological, lexical, syntactic, semantic and situational.
Several studies have comprehensively explored the co-adaptivity in interlocutors on the linguistic structure of the above-mentioned levels.
For example, the empirical results from perception tasks in \citet{pardo2006phonetic} verify the increasing similarity of the phonetic repertoire, which indicates phonetic convergence during conversational interaction.
\citet{garrod1987saying}, in their lab-based study, show that interlocutors in conversational interaction coordinate their utterances to form a mutually acceptable form of description, which indicates the convergence of lexical choice in interaction.

In this paper, we focus on linguistic alignment on the syntactic level.
Our argument is that with the development of mutual understanding during conversational interaction, certain types of syntactic convergence can be observed.
Previous studies found alignment of syntactic complexity, but only for English data, which lacks linguistic generality.
Therefore, we try to find more empirical evidence to show that syntactic alignment happens in other languages, such as German, too.
The goal of this paper is to revisit the syntactic complexity convergence phenomenon discussed by \citet{xu2016convergence} and test whether it holds for German dialogue data, too.
To this end, we selected the following three conversation datasets for German: \textbf{MUNDEX} \cite{TurkWagner2023}, \textbf{TexPrax} \cite{stangier2022texprax}, and \textbf{VERBMOBIL (VM2)} \cite{kay1992verbmobil}.

\section{Background}
\label{sec:related-work}

\subsection{Dependency Structure}

In this paper, we quantify syntactic complexity with the help of dependency parsing \cite{kubler2009dependency}.
We follow the definition of dependency structure by \citet{liu2023linear}.
A linguistic structure, such as a dependency structure, consists of relations of pairs of natural language tokens.
Let $\Sigma$ denote a finite set of natural language tokens (the vocabulary). Let $V = \{w_1, w_2, \ldots ,w_N\}$ denote a spanning node set with its element $w_i \in \Sigma^{\ast}$ \citep{kubler2009dependency}.
The element $w_i$ is a ‘head’ or a dependent in a dependency structure.
The spanning node set $V$ represents a sentence $\omega = w_1w_2\ldots w_N$. 
The dependency structure of the sentence $\omega$ is then a typed structure $\zeta = (V,E,R)$, where $R$ is the set of dependency relation types, $E \subseteq V \times V \times R$ the set of arcs, if $(x,y,r) \in E$, it holds that $\forall r \neq r', (x,y,r') \notin \zeta$.
Under the definition above, a dependency structure is typically a directed acyclic graph (DAG) and the dependency relations within the structure are binary and asymmetric.

We use a statistic and neural sequential model based parsing method, namely the StanfordNLP parser \href{https://stanfordnlp.github.io/stanza/depparse.html}{Stanza} \cite{qi2018universal} for our goal in this paper. Stanza is trained upon the Universal Dependencies (UD) Treebanks \cite{nivre2020universal}.
UD Treebanks store the information about the dependency relations among the lexicon, i.e., given a word, what are the most likely words that can serve as its heads or dependents in a dependency structure.
The core idea can be mathematically expressed as follows based on \citet{zhang-etal-2016-top}:
\begin{equation*}
    P_{\text{head}}(w_j \mid w_i, \mathbf{\vartheta}) = \frac{\exp(g(w_i,w_j))}{\sum_{k=0}^{|\vartheta|}\exp(g(w_i,w_k))}
\end{equation*}
where $\vartheta$ is the lexicon, $g(\cdot)$ is a function which outputs the association score of one word choosing the other word as its head.
$P_{\text{head}}(w_j \mid w_i, \vartheta)$ thus tells us what is the most likely head word $w_j$ given the dependent word $w_i$ and the lexicon. With the generated probability information, the maximum spanning tree algorithm, e.g., Chu-Liu/Edmonds algorithm \citep{chu1965shortest,edmonds1967optimum} is then used to decide what is the most likely dependency structure for a given sentence.

\subsection{Syntactic Complexity}

The topic of syntactic complexity has been of significant interest for researchers working within either functional (cognitive) or computational frameworks of linguistics. According to \citet{szmrecsanyi2004operationalizing}, syntactic complexity refers to syntactic structures which entail increasing cognitive load to parse and process. Sentences that are ranked as more syntactically complex are considered more difficult for humans to process \cite{lin1996structural}.

\citet{szmrecsanyi2004operationalizing} further summarizes three measures for evaluating the syntactic complexity, namely word counts, node counts, and a so-called “Index of Syntactic Complexity”. Word counts use length of a given sentence -- number of words, syllables, intonation units -- to approximate the syntactic complexity, which is based on the straightforward intuition, that a lengthy sentence tends to be more structurally complex than a short one. 
Node count uses the idea that the more phrasal nodes a linguistic unit dominates, the more complex a sentence is \citep[e.g.,][]{rickford1995syntactic}. 
“Index of Syntactic Complexity” focuses on percentage of subordinate clauses \citep{beaman1984coordination} as well as embeddedness of word forms \cite{givon1991markedness}, which is reflected by the following indicators (i) the number subordinating conjunctions, e.g., because, since, etc.; (ii) the number of WH-pronouns, e.g., what, which, etc.; (iii) embeddedness of the verb forms, e.g., finite or infinite; (iv) the number of noun phrases. 

According to \citet{xu2016convergence}, the convergence of syntactic complexity between two speakers in dialogue correlates to two theories: one is the Interactive Alignment theory \citep{pickering2004interactive}, which combines the development of mutual understanding with linguistic alignment.
The other is the Uniform Information Density hypothesis \cite{jaeger2006speakers,jaeger2010redundancy}, which states that speakers will strive to keep information density roughly constant.
Based on this hypothesis, if a speaker decreases its information amount, the other will increase the amount instead. 
According to \citet{jaeger2006speakers} and \citet{jaeger2010redundancy}, information density is expected to be proportional to the complexity of syntactic structure.
This give us an implication that in a dialogue, if a speaker's syntactic complexity is decreasing, the interlocutor's syntactic complexity should be increasing.
This implication is consistent with dependency locality theory \citep[DLT;][]{gibson2000dependency}, which claims that comprehension difficulty is associated with some complex dependency structures. 
The interplay of syntactic complexity and language comprehension has been further investigated in, e.g., \citet{liu2008dependency}, which shows that, average dependency distance positively correlates with the comprehension difficulty (processing effort).

\citet{xu2016convergence} then showed three measures to quantify the syntactic complexity: sentence length, branching factors, and tree depth. Tree depth is used to described how deep a syntactic tree can grow. The deeper a tree is, the more complex a sentence is considered.
Branching factor reports the average number of children of all non-leaf nodes in the parse tree of a sentence. Thus, a syntactic tree that contains, e.g., more constituents or noun phrases within a sentence of a given length, is more complex.

\section{Data}
\label{sec:data}

In order to check the dynamics of syntactic complexity in conversational interaction, we select the following three German datasets for our study:

\textbf{MUNDEX} consists of task-oriented dialogues and focuses on explanation in interaction \cite{TurkWagner2023}.
Each dialogue is a explanation scenario involving a speaker (the explainer) explaining how to play a board game to a recipient (the explainee).
The dataset is still under construction but in total it consists of 87 dialogues between dyads of German native speakers.
At its current stage, speech diarization was mainly performed automatically using Whisper ASR \citep{radford2022robust}.

\textbf{TexPrax} consists of task-oriented dialogues from factory workers on how to solve specific technical issues \cite{stangier2022texprax}. The data are collected anonymously using an open source messaging application in a simulated factory environment. The dataset has in total 202 task-oriented German dialogues containing 1,027 sentences with sentence-level expert annotations, such as turn taking labels. 

The \textbf{VERBMOBIL (VM2)} dataset  \cite{kay1992verbmobil} is based on recordings of various appointment scheduling scenarios, and consists of 30,800 utterances collected in face-to-face interactions.
All utterances are annotated with dialogue acts.

\begin{figure}[ht]
    \centering
    \small
    \textbf{MUNDEX}\\
        \includegraphics[width=0.48\columnwidth]{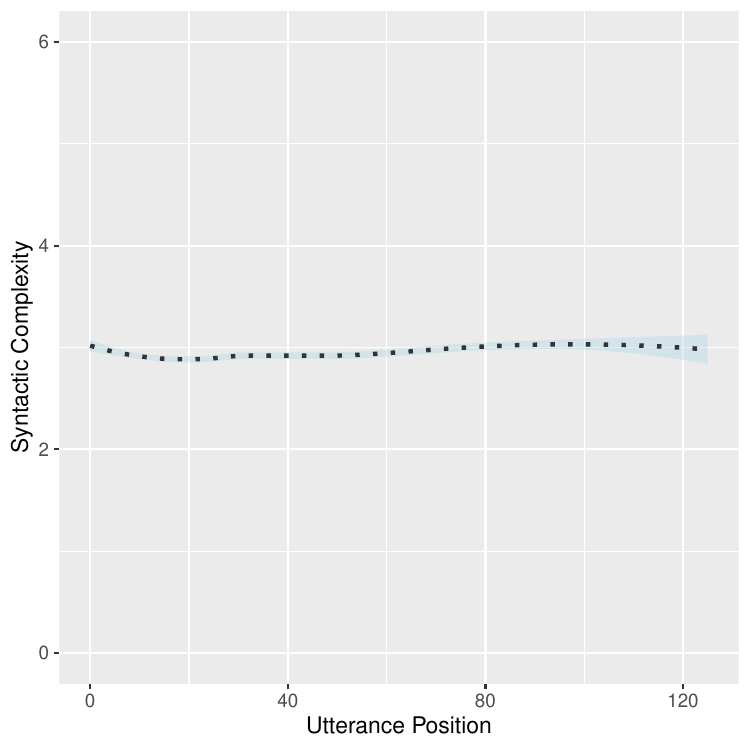}
        \includegraphics[width=0.48\columnwidth]{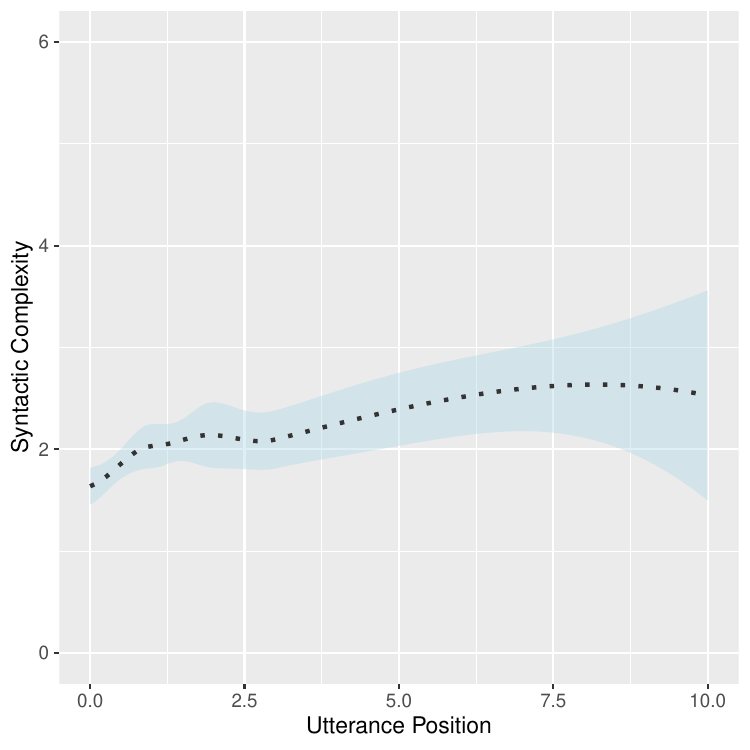}
    \textbf{TexPrax}\\
        \includegraphics[width=0.48\columnwidth]{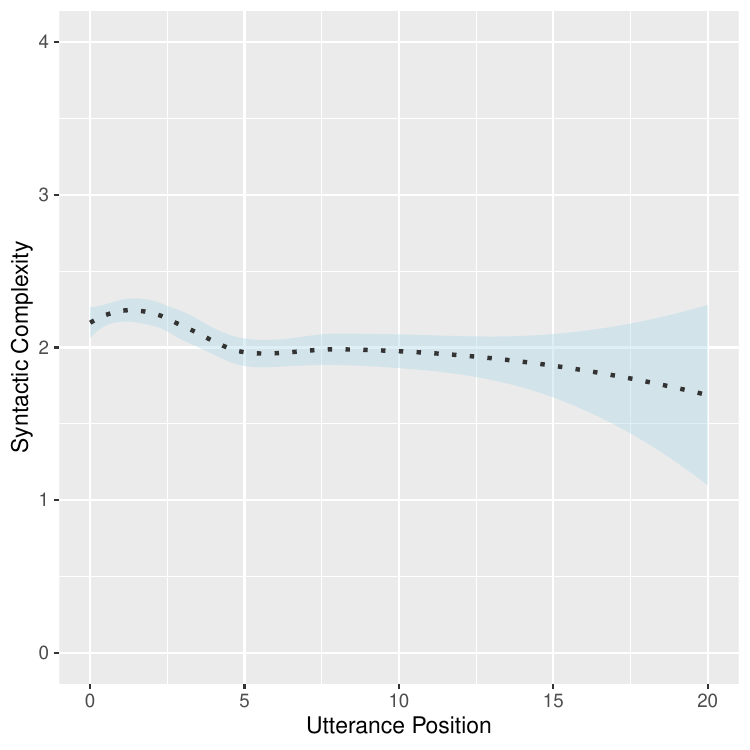}
        \includegraphics[width=0.48\columnwidth]{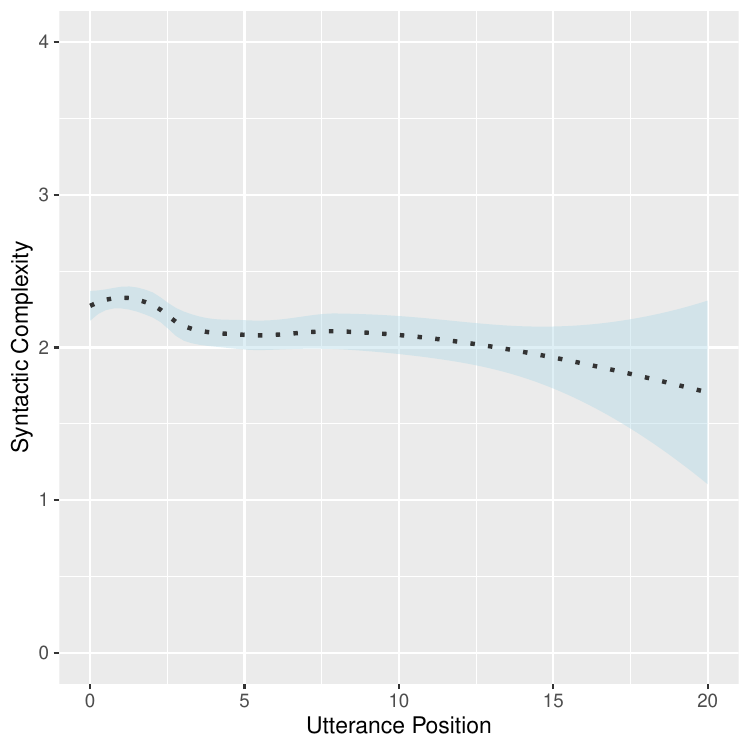}
    \textbf{VM2}\\
        \includegraphics[width=0.48\columnwidth]{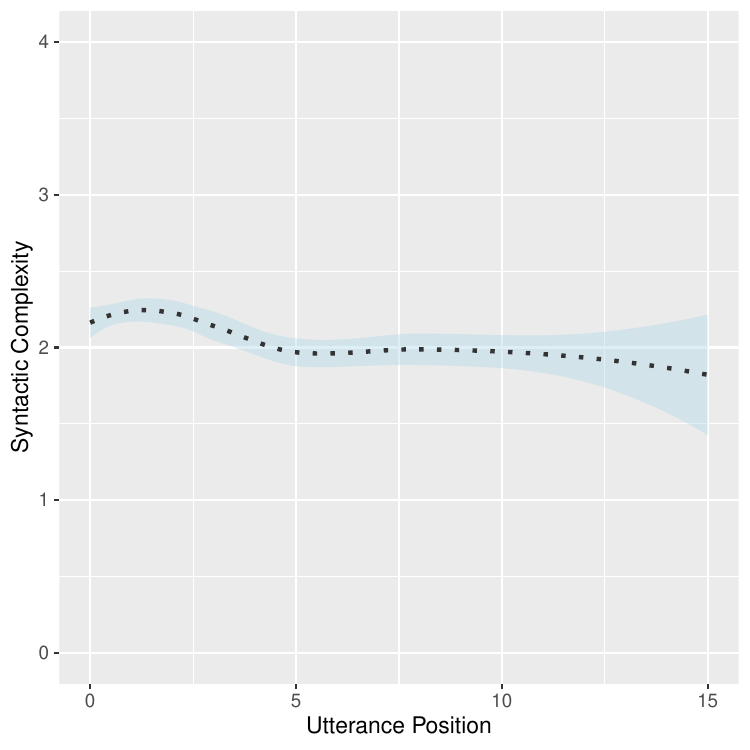}
        \includegraphics[width=0.48\columnwidth]{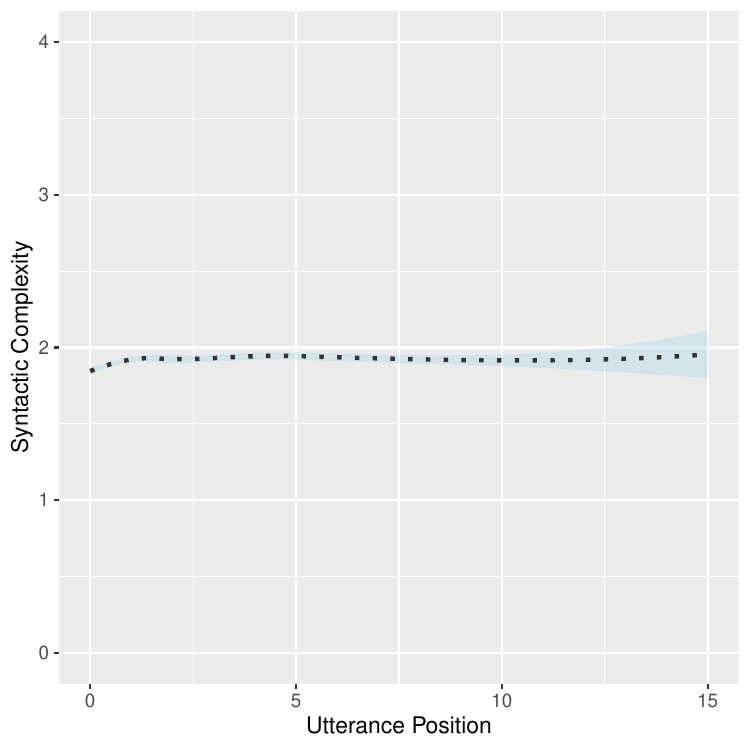}
    \caption{Comparing the development of syntactic complexity of dialogue initiators (left) and followers (right) over the course of the interactions in each corpus. Shaded areas are bootstrapped $95\%$ confidence intervals.}
    \label{fig:syntactic-complexity}
\end{figure}

The main difference among the three datasets is that in  \textbf{MUNDEX}, compared to \textbf{TexPrax} and \textbf{VM2}, one speaker (the explainer) speaks much more than the other (explainee) in every dialogue. This property of the  data has been well reflected in our later analysis (e.g., see Figure~\ref{fig:syntactic-complexity} in Section~\ref{sec:results}). While for the other two datasets, utterance length among the participants is similar.
Moreover, \textbf{VM2} is much larger than the other two selected datasets.

There are two common points among the three selected datasets.
First of all, in each dialogue there are only two dialogue participants.
For the speaker role assignment, we define the interlocutor who initiates the dialogue as \textbf{dialogue initiator}, the other interlocutor who follows the dialogue as \textbf{dialogue follower}.
In this study specifically, we choose to give the role of dialogue initiator to the dialogue participant who starts the conversation.
This is based on our observation in the three datasets that there are no topic shifts in the dialogues.
For example, \textbf{MUNDEX} is based on a pre-defined scenario, where an explainer explains a board game to an explainee. 
Therefore, we do not consider that we need to shift participant roles, as in \citet{xu2016convergence}, which uses the
Switchboard dataset, where each dialogue may have multiple topic shifts.

Secondly, at the end of the interactions, a certain level of mutual understanding can be estimated: in \textbf{MUNDEX}, the explainees are likely to understand the game rules and to be able to play the game; in \textbf{TexPrax}, the workers know the technical issues from their co-worker; in \textbf{VM2} appointments have been successfully made in most of the cases. 
Under this preposition, in this study, by looking at the change of syntactic complexity, namely the phenomenon of syntactic complexity convergence, we assume that we can infer the level of mutual understanding with the development of the dialogue.

\section{Methods}
\label{sec:methods}

To quantify the syntactic complexity, we follow the measures developed in \citet{xu2016convergence}, mainly looking at branching factor, tree depth, and sentence length. Given that all of the three factors can influence the syntactic complexity, it makes sense to quantify the three factors into a single value to represent the syntactic complexity.

We use the number of heads (word count) as a normalisation factor.
In dependency structure, the heads are the nodes which have both incoming and outgoing edges, the tree depths are the maximum number of arcs a tree can have from its root to a terminal node. Given two dependency structures with the same number of heads, if one structure has bigger length, it indicates that the heads in general controls more sub-nodes, and thus the structure is more complex.
Given a speaker's utterance, we calculate utterance length $L$ and use dependency parsing to get the number of heads $\alpha$ as well as the maximum tree depth $\beta$. 
The syntactic complexity $SC$ of the utterance is thus computed as following:
\begin{equation*}
    SC = 
    \begin{cases}
        \lambda \cdot \frac{L}{\alpha} + (1 - \lambda) \cdot \beta  
            & \text{if } \alpha > 0\\
        (1 - \lambda) \cdot \beta  
            & \text{otherwise}
    \end{cases}
\end{equation*}
where $\lambda$ is a tuning factor set to $0.5$ by default.

\begin{figure}
    \centering
    \begin{subfigure}[b]{\columnwidth}
        \centering
        \includegraphics[width=0.75\textwidth]{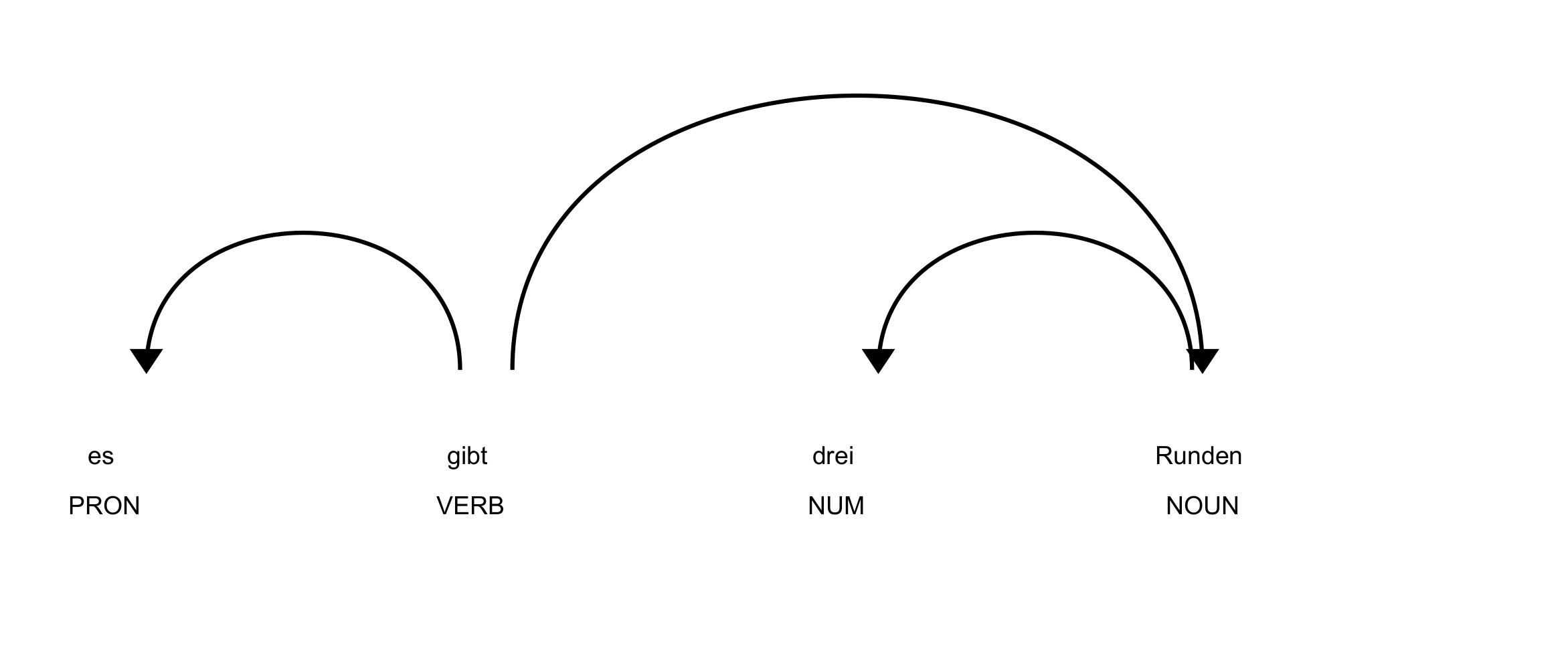}
        \caption{Simple dependency structure (translation: “There are three rounds”).}
         \label{fig:dpsim}
         \vspace{0.5cm}
    \end{subfigure}
    \begin{subfigure}[b]{\columnwidth}
        \centering
        \includegraphics[width=\textwidth]{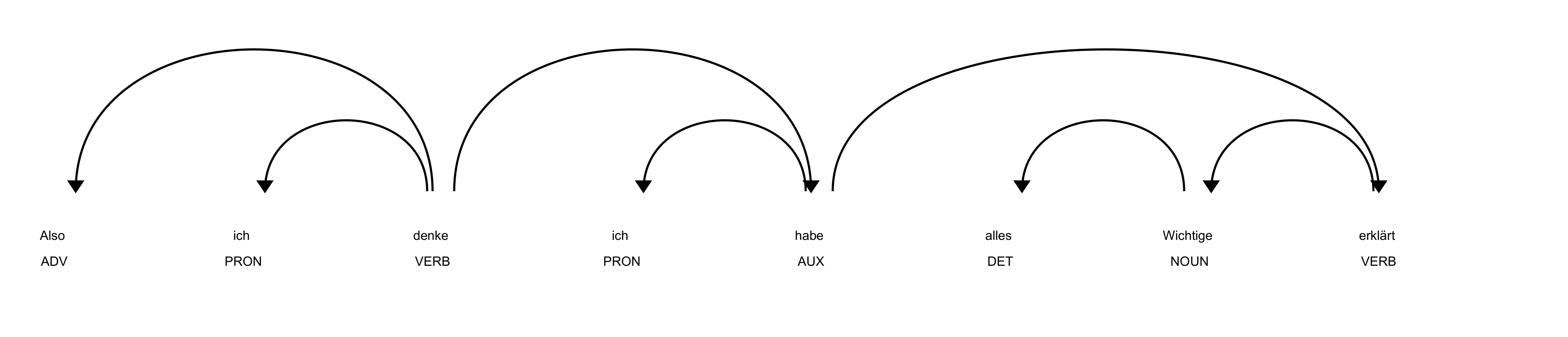}
        \caption{Complex dependency structure (translation: “So I think I have explained all important things”).}
        \label{fig:dpcom}
    \end{subfigure}
    \caption{Two examples showing the dependency structure syntactic relationships according to UD. Edges are directed from heads to dependents.}
    \label{fig:1df}
\end{figure}

Here we use two German example sentences with corresponding dependency trees to show what is considered as syntactically complex. The example in Figure~\ref{fig:dpsim} is a sentence which is considered syntactically simple based on our definition, its maximum tree depth is three and it only has three heads, its sentence length is four.
The example in Figure~\ref{fig:dpcom} in contrast is considered syntactically complex, its maximum tree depth is four and it has four heads, its length is 8.
The quantified syntactic complexity for the first sentence, according to our method, is $2.167$ (three heads as the root node is also considered as a head by Stanza parser, tree depth is three, length is four) while for the second one it is $3$ (four heads as the root node is also considered as a head by Stanza parser, tree depth is four, length is eight).

Moreover, utterances in the three selected datasets have varied length. According to our observation, a speaker may produce multiple utterances before the turn is shifted to a listener, which occurs frequently in the \textbf{MUNDEX} dataset.
Therefore, it is not rational to calculate syntactic complexity values on a turn-by-turn basis. 
As a simple solution, for both dialogue initiator and follower, we calculate the syntactic complexity value on  an utterance-by-utterance basis.
We perform data separation based on the role definition mentioned in in Section~\ref{sec:data}. 

\section{Results and Discussion}
\label{sec:results}

To verify the convergence of syntactic complexity between two speakers in dialogue, we use a linear mixed effects model, specifically regression, to model the dynamics of syntactic complexity (statistics in Table~\ref{tab:statistics}, all reported beta coefficient values are statistically significant).
It turns out that among the three selected datasets, only \textbf{VM2} shows the syntactic complexity convergence, as supported by a negative beta coefficient value for the dialogue initiators and a positive beta coefficient value for the dialogue followers, which indicates that the syntactic complexity of the dialogue initiator generally decreases with the development of the utterance position.
In contrast, the opposite tendency can be observed for the dialogue followers, where the beta coefficient value is positive.

As for the other two selected datasets, in \textbf{MUNDEX}, the beta coefficient value is positive for both dialogue initiators and followers while in \textbf{TexPrax}, the beta coefficient value is instead negative for both dialogue initiators and followers, which indicates that syntactic complexity convergence is not supported by the statistics.

\begin{table}
    \centering
    \small
    \caption{Beta coefficient report on the three dialogue data sets ($^{*}$ represents statistically significant correlations $p < 0.05$, $^{**}$ represents statistically significant correlations $p < 0.01$, and $^{***}$ represents statistically significant correlations $p < 0.001$).}
        \centering
        \begin{tabularx}{\columnwidth}{XXXX}
            \toprule
            & \textbf{MUNDEX} & \textbf{TexPrax} & \textbf{VM2} \\
            \midrule
            \textbf{Initiator} & $0.0009^{**}$ & $-0.02^{***}$ & $-0.02^{***}$ \\
            \textbf{Follower}  & $0.14^{***}$  & $-0.22^{***}$ &  $0.005^*$\\
            \bottomrule
        \end{tabularx}
        \label{tab:statistics}
\end{table}

\begin{table}
\small
\caption{Range of syntactic complexity values for dialogue initiators and followers across corpora.}
\label{tab:sc-range}
\begin{tabularx}{\columnwidth}{lXXXX}
\toprule
            & \multicolumn{2}{l}{\textbf{SC Initiators}} & \multicolumn{2}{l}{\textbf{SC Followers}} \\
            \cmidrule{2-3} \cmidrule{4-5}
            & min & max & min & max \\
\midrule
    \textbf{MUNDEX}  & 1.5 & 5.9 & 1 & 3.2 \\
    \textbf{TexPrax} & 1 & 4.73 & 1 & 4.14 \\
    \textbf{VM2}     & 1 & 4.14 & 1 & 4.57 \\
\bottomrule
\end{tabularx}
\end{table}

Looking at the plots in Figure~\ref{fig:syntactic-complexity}, it seems that the increasing/decreasing tendencies are small but still obvious in \textbf{VM2}. 
This can be explained, at least in part, by the relatively small values of the beta coefficients. Nevertheless, given that the range of syntactic complexity values is not so large (see Table~\ref{tab:sc-range}), we assume that the reported effect sizes are valid.
For the \textbf{MUNDEX} dataset, it turns out that dialogue followers' syntactic complexity is gradually increasing, while dialogue initiators' syntactic complexity remains quite stable, although it is slightly increasing as well.
We considered this as a different type of syntactic complexity convergence.
One possible explanations could be that, in \textbf{MUNDEX}'s scenario, the explainers have to continuously introduce different rules and constraints of the game, and thus the syntactic complexity value for dialogue initiators slightly increased (as evidenced by the statistics in Table~\ref{tab:statistics}). While for the dialogue followers, with the development of an explanation, they got more engaged and thus started to use more complex structures or produce longer utterances.
In the \textbf{TexPrax} dataset, a general decreasing trend can be observed for both dialogue initiators and followers, which is in general not consistent with the phenomenon of syntactic complexity convergence.

From an information-theoretic perspective, the convergence of syntactic complexity between dialogue participants reflects the convergence of shared information \cite{genzel2002entropy,genzel2003variation}, which is seen as evidence that dialogue participants are working co-constructively to build common ground \cite{clark1996using}.
The results reported in this study show that the convergence of syntactic complexity as a linguistic phenomenon can be observed in dialogues, (1) in different languages (e.g., in English and at least partially in German); (2) under different scenarios (e.g., explaining a game in \textbf{MUNDEX} or making an appointment in \textbf{VM2}).

\section{Conclusions}
\label{sec:conclusions}

In this paper, we revisit the phenomenon of syntactic complexity convergence by examining it specifically for German dialogue data.
The convergence of syntactic complexity is assumed to be strongly related to the uniform information density theory as well as to the interactive alignment theory, which correlates the development of mutual understanding with linguistic alignment.
Our empirical results show that the convergence also exists in one of the three German dialogue datasets we analysed, which provides further evidence for the generality of syntactic complexity convergence.
Given that the \textbf{VM2} dataset is much larger than the other two datasets, we are prone to claiming that syntactic complexity convergence has its linguistic generality.
We also found a different type of syntactic complexity convergence in the \textbf{MUNDEX} dataset, while further investigation is still necessary.

\section*{Acknowledgements}
This research was funded by the \href{https://www.dfg.de}{Deutsche For\-schungs\-gemeinschaft (DFG)}: \href{https://gepris.dfg.de/gepris/projekt/438445824}{TRR 318/1 2021 – 438445824}.

\section*{Limitations}

When processing German utterances, we did not consider possible solutions to deal with disfluencies.
One possible solution would have been to replace disfluent sentences with fluent (i.e., grammatical) ones. This, however, could change the syntactic complexity values. In order to take into account the effect of disfluencies on syntactic complexity, an empirical study on whether disfluencies increases syntactic complexity needs to be carried out beforehand.
Another issue we haven't explored further is whether linear models are optimal for our data analysis. A potential future work is to fit a model with a quadratic term for hypothesis testing.

\section*{Ethics statement}
Given the scope of this study, there do not appear to be any ethical issues.

\bibliography{konvens2024}

\begin{thebibliography}{27}
\expandafter\ifx\csname natexlab\endcsname\relax\def\natexlab#1{#1}\fi

\bibitem[{Beaman(1984)}]{beaman1984coordination}
Karen Beaman. 1984.
\newblock Coordination and subordination revisited: Syntactic complexity in spoken and written narrative discourse.
\newblock In Deborah Tannen, editor, \emph{Spoken and Written Language. Exploring Orality and Literacy}, pages 45--80. Ablex.

\bibitem[{Chu(1965)}]{chu1965shortest}
Yoeng-Jin Chu. 1965.
\newblock On the shortest arborescence of a directed graph.
\newblock \emph{Scientia Sinica}, 14:1396--1400.

\bibitem[{Clark(1996)}]{clark1996using}
Herbert~H. Clark. 1996.
\newblock \href {https://doi.org/10.1017/CBO9780511620539} {\emph{Using Language}}.
\newblock Cambridge University Press, Cambridge, UK.

\bibitem[{Edmonds et~al.(1967)}]{edmonds1967optimum}
Jack Edmonds et~al. 1967.
\newblock Optimum branchings.
\newblock \emph{Journal of Research of the National Bureau of Standards B}, 71(4):233--240.

\bibitem[{Garrod and Anderson(1987)}]{garrod1987saying}
Simon Garrod and Anthony Anderson. 1987.
\newblock \href {https://doi.org/10.1016/0010-0277(87)90018-7} {Saying what you mean in dialogue: A study in conceptual and semantic co-ordination}.
\newblock \emph{Cognition}, 27:181--218.

\bibitem[{Genzel and Charniak(2002)}]{genzel2002entropy}
Dmitriy Genzel and Eugene Charniak. 2002.
\newblock \href {https://doi.org/10.3115/1073083.1073117} {Entropy rate constancy in text}.
\newblock In \emph{Proceedings of the 40th Annual Meeting of the Association for Computational Linguistics}, pages 199--206, Philadelphia, PA, USA.

\bibitem[{Genzel and Charniak(2003)}]{genzel2003variation}
Dmitriy Genzel and Eugene Charniak. 2003.
\newblock \href {https://aclanthology.org/W03-1009} {Variation of entropy and parse trees of sentences as a function of the sentence number}.
\newblock In \emph{Proceedings of the 2003 Conference on Empirical Methods in Natural Language Processing}, pages 65--72.

\bibitem[{Gibson(2000)}]{gibson2000dependency}
Edward Gibson. 2000.
\newblock The dependency locality theory: A distance-based theory of linguistic complexity.
\newblock In \emph{Image, Language, Brain: Papers from the First Mind Articulation Project Symposium}, pages 95--126. The MIT Press.

\bibitem[{Giv{\'o}n(1991)}]{givon1991markedness}
Talmy Giv{\'o}n. 1991.
\newblock \href {https://doi.org/10.1075/sl.15.2.05giv} {Markedness in grammar: Distributional, communicative and cognitive correlates of syntactic structure}.
\newblock \emph{Studies in Language}, 15(2):335--370.

\bibitem[{Jaeger(2010)}]{jaeger2010redundancy}
T.~Florian Jaeger. 2010.
\newblock \href {https://doi.org/10.1016/j.cogpsych.2010.02.002} {Redundancy and reduction: Speakers manage syntactic information density}.
\newblock \emph{Cognitive Psychology}, 61:23--62.

\bibitem[{Jaeger and Levy(2006)}]{jaeger2006speakers}
T.~Florian Jaeger and Roger Levy. 2006.
\newblock \href {https://proceedings.neurips.cc/paper_files/paper/2006/hash/c6a01432c8138d46ba39957a8250e027-Abstract.html} {Speakers optimize information density through syntactic reduction}.
\newblock \emph{Advances in Neural Information Processing Systems}, 19.

\bibitem[{Kay(1992)}]{kay1992verbmobil}
Martin Kay. 1992.
\newblock \emph{Verbmobil: A Translation System for Face-to-Face Dialog}.
\newblock University of Chicago Press.

\bibitem[{K{\"u}bler et~al.(2009)K{\"u}bler, McDonald, and Nivre}]{kubler2009dependency}
Sandra K{\"u}bler, Ryan McDonald, and Joakim Nivre. 2009.
\newblock \href {https://doi.org/10.1007/978-3-031-02131-2_2} {\emph{Dependency parsing}}, pages 11--20. Springer.

\bibitem[{Lin(1996)}]{lin1996structural}
Dekang Lin. 1996.
\newblock \href {https://aclanthology.org/C96-2123} {On the structural complexity of natural language sentences}.
\newblock In \emph{{COLING} 1996 Volume 2: The 16th International Conference on Computational Linguistics}, pages 729--733, Copenhagen, Denmark.

\bibitem[{Liu(2008)}]{liu2008dependency}
Haitao Liu. 2008.
\newblock Dependency distance as a metric of language comprehension difficulty.
\newblock \emph{Journal of Cognitive Science}, 9(2):159--191.

\bibitem[{Liu et~al.(2023)Liu, Amini, Sachan, and Cotterell}]{liu2023linear}
Tianyu Liu, Afra Amini, Mrinmaya Sachan, and Ryan Cotterell. 2023.
\newblock \href {https://doi.org/10.18653/v1/2023.emnlp-main.52} {Linear-time modeling of linguistic structure: An order-theoretic perspective}.
\newblock In \emph{Proceedings of the 2023 Conference on Empirical Methods in Natural Language Processing}, pages 808--830, Singapore.

\bibitem[{Nivre et~al.(2020)Nivre, de~Marneffe, Ginter, Haji{\v{c}}, Manning, Pyysalo, Schuster, Tyers, and Zeman}]{nivre2020universal}
Joakim Nivre, Marie-Catherine de~Marneffe, Filip Ginter, Jan Haji{\v{c}}, Christopher~D. Manning, Sampo Pyysalo, Sebastian Schuster, Francis Tyers, and Daniel Zeman. 2020.
\newblock \href {https://aclanthology.org/2020.lrec-1.497} {{U}niversal {D}ependencies v2: An evergrowing multilingual treebank collection}.
\newblock In \emph{Proceedings of the Twelfth Language Resources and Evaluation Conference}, pages 4034--4043, Marseille, France.

\bibitem[{Pardo(2006)}]{pardo2006phonetic}
Jennifer~S. Pardo. 2006.
\newblock \href {https://doi.org/10.1121/1.2178720} {On phonetic convergence during conversational interaction}.
\newblock \emph{The Journal of the Acoustical Society of America}, 119:2382--2393.

\bibitem[{Pickering and Garrod(2004)}]{pickering2004interactive}
Martin~J. Pickering and Simon Garrod. 2004.
\newblock \href {https://doi.org/10.1017/S0140525X04000056} {The interactive-alignment model: Developments and refinements}.
\newblock \emph{Behavioral and Brain Sciences}, 27(2):212--225.

\bibitem[{Qi et~al.(2018)Qi, Dozat, Zhang, and Manning}]{qi2018universal}
Peng Qi, Timothy Dozat, Yuhao Zhang, and Christopher~D. Manning. 2018.
\newblock \href {https://doi.org/10.18653/v1/K18-2016} {{U}niversal {D}ependency parsing from scratch}.
\newblock In \emph{Proceedings of the {C}o{NLL} 2018 Shared Task: Multilingual Parsing from Raw Text to Universal Dependencies}, pages 160--170, Brussels, Belgium.

\bibitem[{Radford et~al.(2022)Radford, Kim, Xu, Brockman, McLeavey, and Sutskever}]{radford2022robust}
Alec Radford, Jong~Wook Kim, Tao Xu, Greg Brockman, Christine McLeavey, and Ilya Sutskever. 2022.
\newblock \href {https://doi.org/10.48550/arXiv.2212.04356} {Robust speech recognition via large-scale weak supervision}.
\newblock \emph{Audio and Speech Processing (eess.AS)}, arXiv:2212.04356.

\bibitem[{Rickford and Wasow(1995)}]{rickford1995syntactic}
John~R. Rickford and Thomas~A. Wasow. 1995.
\newblock \href {https://doi.org/10.2307/415964} {Syntactic variation and change in progress: Loss of the verbal coda in topic-restricting as far as constructions}.
\newblock \emph{Language}, 71:102--131.

\bibitem[{Stangier et~al.(2022)Stangier, Lee, Wang, M{\"u}ller, Frick, Metternich, and Gurevych}]{stangier2022texprax}
Lorenz Stangier, Ji-Ung Lee, Yuxi Wang, Marvin M{\"u}ller, Nicholas Frick, Joachim Metternich, and Iryna Gurevych. 2022.
\newblock \href {https://aclanthology.org/2022.aacl-demo.2} {{T}ex{P}rax: A messaging application for ethical, real-time data collection and annotation}.
\newblock In \emph{Proceedings of the 2nd Conference of the Asia-Pacific Chapter of the Association for Computational Linguistics and the 12th International Joint Conference on Natural Language Processing: System Demonstrations}, pages 9--16, Taipei, Taiwan.

\bibitem[{Szmrecsányi(2004)}]{szmrecsanyi2004operationalizing}
Benedikt~M. Szmrecsányi. 2004.
\newblock On operationalizing syntactic complexity.
\newblock In \emph{Le poids des mots. Proceedings of the 7th International Conference on Textual Data Statistical Analysis}, volume~2, pages 1031--1038, Louvain-la-Neuve, Belgium.

\bibitem[{Türk et~al.(2023)Türk, Wagner, Buschmeier, Grimminger, Wang, and Lazarov}]{TurkWagner2023}
Olcay Türk, Petra Wagner, Hendrik Buschmeier, Angela Grimminger, Yu~Wang, and Stefan Lazarov. 2023.
\newblock \href {http://nbn-resolving.de/urn:nbn:de:0070-pub-29805458} {Mundex: A multimodal corpus for the study of the understanding of explanations}.
\newblock In \emph{Proceedings of the 1st International Multimodal Communication Symposium}, pages 63--64, Barcelona, Spain.

\bibitem[{Xu and Reitter(2016)}]{xu2016convergence}
Yang Xu and David Reitter. 2016.
\newblock \href {https://doi.org/10.18653/v1/P16-2072} {Convergence of syntactic complexity in conversation}.
\newblock In \emph{Proceedings of the 54th Annual Meeting of the Association for Computational Linguistics (Volume 2: Short Papers)}, pages 443--448, Berlin, Germany.

\bibitem[{Zhang et~al.(2016)Zhang, Lu, and Lapata}]{zhang-etal-2016-top}
Xingxing Zhang, Liang Lu, and Mirella Lapata. 2016.
\newblock \href {https://doi.org/10.18653/v1/N16-1035} {Top-down tree long short-term memory networks}.
\newblock In \emph{Proceedings of the 2016 Conference of the North {A}merican Chapter of the Association for Computational Linguistics: Human Language Technologies}, pages 310--320, San Diego, California. Association for Computational Linguistics.

\end{thebibliography}
\bibliographystyle{acl_natbib}

\end{document}